\documentclass{INTERSPEECH2023}

\usepackage{xcolor}
\usepackage{amsmath,graphicx, microtype}
\usepackage{booktabs}
\usepackage{hyperref}
\usepackage{caption}
\usepackage{subcaption}
\usepackage{booktabs}
\usepackage{tabularx}
\usepackage{adjustbox}

\graphicspath{ {./images/} }

\interspeechcameraready

\title{The ART of Conversation: Measuring Phonetic Convergence and \\Deliberate Imitation in L2-Speech with a Siamese RNN}
\name{Zheng Yuan$^{1,2}$, Aldo Pastore$^{1,2}$, Dorina de Jong$^{1,2}$, Hao Xu$^3$, Luciano Fadiga$^{1,2}$,\\ Alessandro D'Ausilio$^{1,2}$}

\address{
  $^1$Istituto Italiano di Tecnologia, Italy \\
  $^2$Università degli Studi di Ferrara, Italy\\
  $^3$ University of California San Diego, USA}

\email{zheng.yuan@iit.it}

\begin{document}

\maketitle
 
\begin{abstract}
Phonetic convergence describes the automatic and unconscious speech adaptation of two interlocutors in a conversation. This paper proposes a Siamese recurrent neural network (RNN) architecture to measure the convergence of the holistic spectral characteristics of speech sounds in an L2-L2 interaction. We extend an alternating reading task (the ART) dataset by adding 20 native Slovak L2 English speakers. We train and test the Siamese RNN model to measure phonetic convergence of L2 English speech from three different native language groups: Italian (9 dyads), French (10 dyads) and Slovak (10 dyads). Our results indicate that the Siamese RNN model effectively captures the dynamics of phonetic convergence and the speaker's imitation ability. Moreover, this text-independent model is scalable and capable of handling L1-induced speaker variability.
\end{abstract}
\noindent\textbf{Index Terms}: phonetic convergence, Siamese RNN, speech imitation, alternating reading task, L2 English

\section{Introduction}
\label{sec:intro}

Phonetic convergence is a phenomenon in which speakers modify their acoustic-phonetic repertoire to approximate that of their conversational partners  \cite{pardo2006phonetic} for various purposes such as enhancing social affiliation, facilitating communication \cite{giles1991accommodation, pickering2004toward}, expressing identity or improving language proficiency \cite{lewandowski2019phonetic}, sometimes unconsciously or without a clear intention \cite{pardo2013measuring}.

Phonetic convergence also reflects the complex interactions between social and cognitive factors that influence speech perception and production \cite{pickering2004toward, goldinger1998echoes, liberman2000relation}. It can reveal how speakers store and process linguistic representations in the brain, how they monitor their own and others' speech, and how they use social cues to guide their linguistic behaviour  \cite{pardo2012reflections, gambi2013prediction}. It also sheds light on the mechanisms of sound change and language evolution \cite{pardo2017phonetic}.

In the context of language acquisition, phonetic convergence can affect L2 development positively by allowing learners to adjust to native-like pronunciation and prosody or negatively by strengthening their L1 features or non-native features adopted from other L2 speakers \cite{chang2019phonetics, olson2019feature, gnevsheva2021phonetic}. Despite extensive research on phonetic convergence involving native speakers \cite{pardo2017phonetic, wagner2021phonetic, jiang2022impact}, the question of whether such convergence also occurs between non-native speakers remains largely unanswered \cite{olmstead2021phonetic}. To study the dynamics of phonetic convergence in L2-L2 interaction, we designed the Alternating Reading Task (ART)\cite{dejong22_interspeech}, a scripted text reading experiment, based on previous studies\cite{bailly2010speech, bailly2014assessing, mukherjee2017relationship, mukherjee2018analyzing, mukherjee2019neural, aubanel2020speaking}. ART maintains the turn-taking structure in natural conversations and provides a controllable experimental complexity. According to past research, a speaker's ability to imitate can be an important factor for phonetic convergence (implicit imitation) \cite{goldinger1998echoes, gambi2013prediction,  garnier2013neural}, we incorporated an explicit imitation condition in our experiment (see Section \ref{sec:ART}). The ART dataset that we collected comprised L2 English speech data from native speakers of Italian, French, and Slovak.

How to measure the degree of phonetic convergence remains an open area of research, with work featuring both subjective evaluations \cite{pardo2006phonetic, goldinger1998echoes, babel2012role} and objective modelling \cite{mukherjee2017relationship, aubanel2020speaking, Levitan2011Measuring}. In this work, we focus on modelling the holistic convergence as it is a more direct indicator of the interaction between speech perception and production \cite{babel2012role}. It is measured by globally comparing temporal and spectral characteristics of two speech signals\cite{delvaux2007influence}. The classic speaker verification method Gaussian Mixture Model-Universal Background Model (GMM-UBM) \cite{reynolds1997comparison} proved robust in assessing holistic phonetic convergence at the word\cite{bailly2014assessing, mukherjee2017relationship, mukherjee2018analyzing, mukherjee2019neural} and sentence level \cite{dejong22_interspeech} by comparing the log-likelihood ratio (LLR) of a speaker's baseline and interactive condition. However, GMM-UBM has its limitations when the speaker number grows large with increasing dialect, language and speaker variability. In this paper, we present a Siamese neural network  \cite{bromley1993signature} with a recurrent architecture (Siamese RNN) \cite{mueller2016siamese}, to measure phonetic convergence and speech similarity \footnote{The code is available at \url{https://github.com/byronthecoder/S-RNN-4-ART}.}, exploiting the neural network's capacity to learn intricate and non-linear speaker feature representations  \cite{snyder2017deep}. To our best knowledge, Siamese neural networks have not been adopted and experimented on phonetic convergence studies.

\section{Alternating Reading Task dataset}
\label{sec:ART}
\subsection{Participants}
\label{ssec:participants}
We recruited 58 participants for the ART experiment: 20 native French (all female, average age 23.45±4.94), 18 native Italian (6 males, average age 24.50±3.65) in the initial version\cite{de_jong_2021_dataset}, and 20 native Slovak (10 males, average age 33.75±13.69) for this phase\cite{zheng_yuan_2023_7993783}. To ensure B2-level English reading competency, all participants passed an online proficiency test \cite{lemhofer2012introducing} (test score: French=82.59±9.53, Italian=74.16±6.70, Slovak=78.12±10.24). Same-sex dyads were formed with participants having similar reading proficiency (\textless $15\%$ difference in test scores), and most of them did not know each other prior to the experiment.

\subsection{Task description}
\label{ssec:task}
\begin{figure}[t]
    \centering
    \includegraphics[width=\columnwidth]{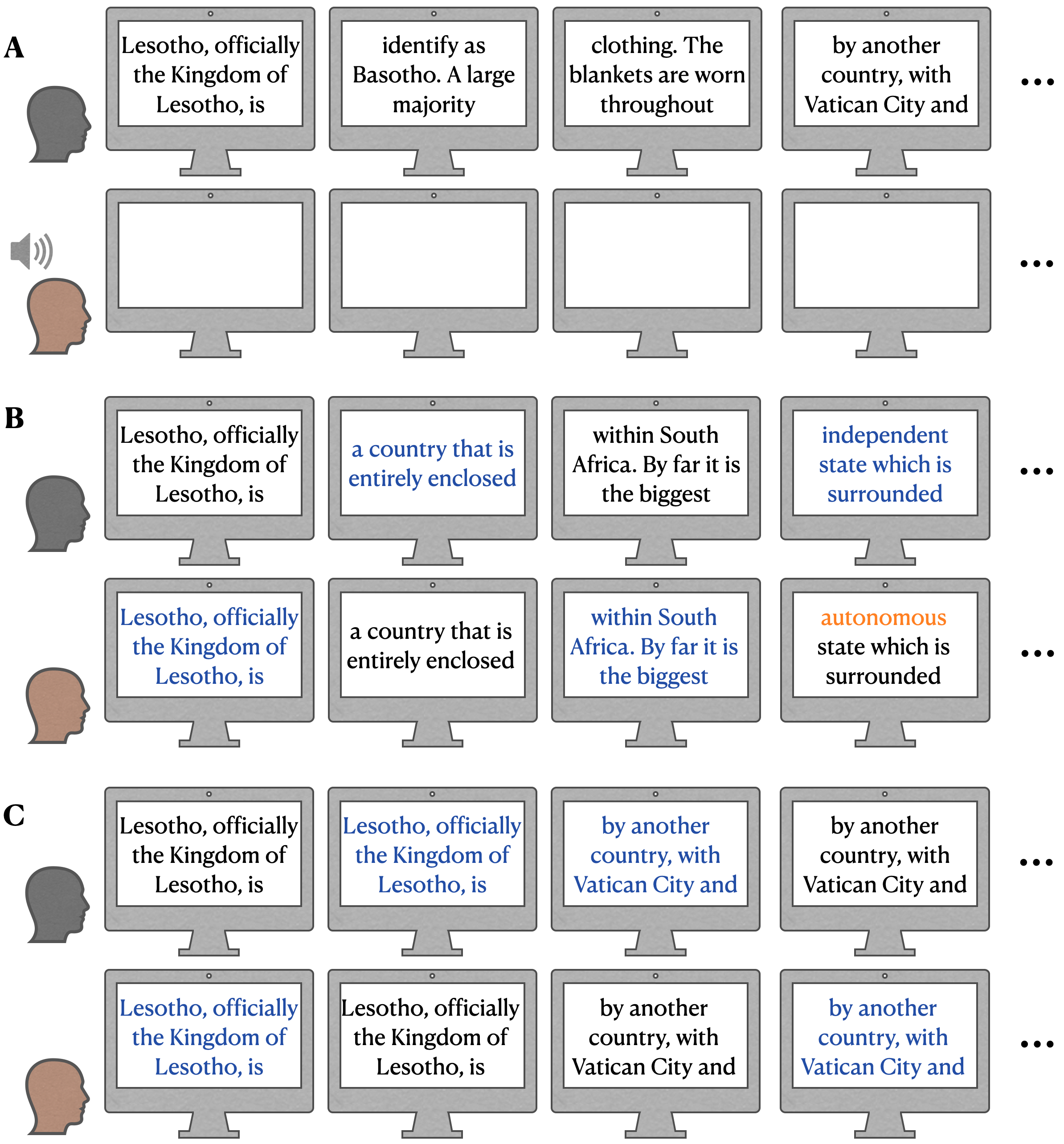}
    \caption{The Alternating Reading Task. Participants speak when their computer screen shows black sentences. (A) The solo condition; (B) the interactive condition - synonym shown in orange for illustrative purpose only, and (C) the imitation condition.}
    \label{fig:ART_demo}
\end{figure}

In the ART experiment\cite{dejong22_interspeech}, dyads took turns reading aloud a neutral English text. The text contained 80 sentences and turn boundaries were set within sentences. We replaced some words with synonyms to maintain attention. To make comparisons, a solo and an imitation condition were added. In the solo session, participants read the sentences individually, and in the interactive condition, participants performed the Alternating Reading Task four times with varying degrees of word replacement. The final imitation session required participants to imitate each other during sentence reading. Participants could not see each other during the experiment (see Figure \ref{fig:ART_demo}).

\section{Siamese RNN model}
\label{sec: model}
We implemented a Siamese recurrent neural network based on Mueller and Thyagarajan's paper \cite{mueller2016siamese} to predict speech utterance similarity and capture speaker information via a binary speaker verification task (see Section \ref{sec:spk_vrf}). This algorithm learns a vector representation for each utterance that encodes the acoustic characteristics of the speaker, allowing for direct measurement of phonetic convergence between interlocutors. Additionally, the RNN's ability to preserve historical information during computation leads to refined modelling of speech effects like coarticulation.

\begin{figure}[h]
    \centering
    \includegraphics[width=\columnwidth]{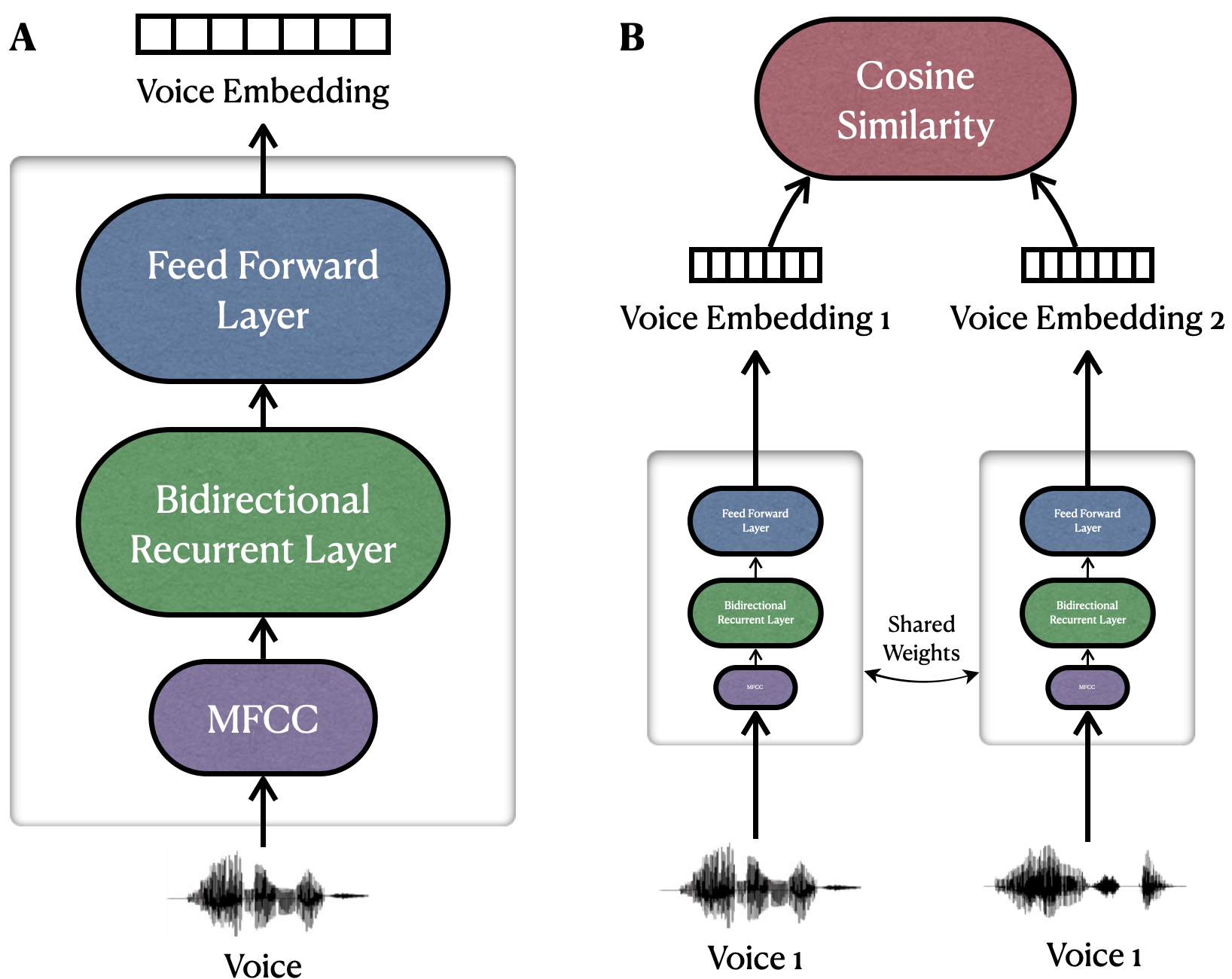}
    \caption{(A) The proposed voice representation module consists of an MFCC extraction module (purple), a bi-directional RNN layer (green), and a feed-forward layer (blue). It generates a low-dimensional vector representation for input voice audio.
    (B) shows our model pipeline. The model inputs two voice audios and computes the distance between their voice embeddings from two tied-weight voice representation modules.}
    \label{fig:siamese_net}
\end{figure}

As shown in Figure \ref{fig:siamese_net}
, we employed a two-network structure, with each RNN network processing one audio input. 
The structure utilizes tied weights, meaning that the trainable parameters in $RNN_1$ and $RNN_2$ are shared. 
Each input speech sentence $s \in S$ consists of a set of fixed length vector $\{\mathbf{x}^s_t\}_T$, where $S$ denotes the corpus, and time step $t\in\{1,...,T\}$, obtained from MFCC extraction (see \ref{ssec:dataset}). The network maps the input vector of dimension $d_{in}$ to a representation space of dimension $d_{rep}$ ($d_{in}=39$ and $d_{rep}=50$ in our experiments). 

Going into the bi-directional RNN layer, the input vector $\mathbf{x}_{t}$ at time $t$ is used to compute both the forward and backwards hidden states $\overrightarrow{\mathbf{h}}_t$ and $\overleftarrow{\mathbf{h}}_t$. The computation for each sentence $s\in S$ is as follows:

\begin{align}
    \overrightarrow{\mathbf{h}}^s_t &= \tanh(\overrightarrow{\mathbf{W}}_h \mathbf{x}^s_t + \overrightarrow{\mathbf{U}}_h \overrightarrow{\mathbf{h}}^s_{t-1} + \overrightarrow{\mathbf{b}}_h), \\
    \overleftarrow{\mathbf{h}}^s_t &= \tanh(\overleftarrow{\mathbf{W}}_h \mathbf{x}^s_t + \overleftarrow{\mathbf{U}}_h \overleftarrow{\mathbf{h}}^s_{t+1} + \overleftarrow{\mathbf{b}}_h), 
\end{align}

\noindent where $\mathbf{W}_h$ and $\mathbf{U}_h$ are weight matrices for all speech segment vectors $\{\mathbf{x}_t\}_T$ and hidden-state vectors $\mathbf{h}_t$ respectively, and $\mathbf{b}_h$ denotes the bias. The hidden-state vectors are initiated as zero vectors $\overrightarrow{\mathbf{h}}_1^s = \overleftarrow{\mathbf{h}}_T^s = \mathbf{0}$. Next, we concatenate the forward and backwards hidden states to obtain the hidden state $\mathbf{h}^s_{t} = [\overrightarrow{\mathbf{h}}^s, \overleftarrow{\mathbf{h}}^s] $, and fed it into a feedforward layer activated by the $\tanh$ function:

\begin{equation}
    \mathbf{y}^s_t = \tanh(\mathbf{W}_y \mathbf{h}^s_t + \mathbf{b}_y),
\end{equation}
\noindent 


\noindent where $\mathbf{W}_{y}$ and $\mathbf{b}_{y}$ are the weight matrix and the bias vector associated with the output $\mathbf{y}^s_t$. 
Additionally, a masking layer has been set before the bi-directional RNNs to cope with sequences of different lengths.

After processing the last time step $T$, the output on sentence $s$ is $\mathbf{y}_T^s$. Then, the output
will go through a feedforward layer activated by the sigmoid function and get transformed into the speech embedding vector $\mathbf{e}^s$ via

\begin{equation}
    \mathbf{e}^s = \text{sigmoid} (\mathbf{W}_e \mathbf{y}_T^s + \mathbf{b}_e),
\end{equation}
\noindent where the weights and bias are denoted by $\mathbf{W}_e$ and $\mathbf{b}_e$.  

The final step is computing the distances between two speech embedding vectors. In this study, we use cosine similarity to measure the distance.
Given the speech embeddings $\mathbf{e}^{s}$ and $\mathbf{e}^{s'}$ for sentence $s,s'\in S$, we compute

\begin{equation}
    g(\mathbf{e}^{s}, \mathbf{e}^{s'}) = \frac{\mathbf{e}^{s} \cdot \mathbf{e}^{s'}}{|\mathbf{e}^{s}|_2 |\mathbf{e}^{s'}|_2}
\end{equation}





\section{Speaker verification task}
\label{sec:spk_vrf}

\newcolumntype{C}{@{\hspace{0.1cm}}c@{\hspace{0.1cm}}}
\begin{table*}[t]
\caption{Model performance on different architectures, training data size and test scenarios} 
\begin{subtable}[t]{0.36\textwidth}
\centering
\begin{tabularx}{\textwidth}{@{}|lCCCC@{}|}
\toprule
Model & Epoch & Size(K) & Param(K) & Acc \\
\midrule
LSTM+FF & 10 & 91.6 & \textbf{20.7} & 0.79 \\
RNN+FF & 10 & 91.6 & 7.2 & 0.66 \\
Bi-RNN+FF & 10 & 91.6 & 14.3 & \textbf{0.84} \\
\bottomrule
\end{tabularx}
\caption{}
\end{subtable}\hspace{0.1cm}
\begin{subtable}[t]{0.23\textwidth}
\centering
\begin{tabularx}{\textwidth}{@{}lCCC@{}|}
\toprule
Model & Epoch & Size(K) & Acc \\
\midrule
VIFS10 & 20 & 4.2 & 0.80 \\
VIFS10 & 100 & 4.2 & \textbf{0.87} \\
VIFS10 & 150 & 4.2 & \textbf{0.87} \\
\bottomrule
\end{tabularx}
\caption{}
\end{subtable}\hspace{0.1cm}
\begin{subtable}[t]{0.15\textwidth}
\centering
\begin{tabularx}{\textwidth}{@{}lCC@{}|}
\toprule
Model & Subset & Acc \\
\midrule
VCTK & IT & 0.70 \\
VCTK & FR & \textbf{0.82} \\
VCTK & SK & 0.75 \\
\bottomrule
\end{tabularx}
\caption{}
\end{subtable}\hspace{0.1cm}
\begin{subtable}[t]{0.22\textwidth}
\centering
\begin{tabularx}{\textwidth}{@{}lCCC@{}|}
\toprule
Model & Size(K) & Epoch & Acc \\
\midrule
IF & 31.6 & 10 & 0.85 \\
\\
IFS & 31.6 & 10 & \textbf{0.90} \\
\bottomrule
\end{tabularx}
\caption{}
\end{subtable}
\label{table:1}
\end{table*}

To capture convergence, we trained a Siamese RNN model using a binary speaker verification task. The model predicts whether two speech utterances are produced by the same speaker based on a cosine similarity score. In our ART dataset, the same sentences were produced across the three experimental conditions. We assume that there is a gradual increase in convergence from the solo to the interactive and imitation conditions. A correct prediction that a sentence pair is produced by different speakers, along with an increased similarity score, indicates phonetic convergence. Likewise, speakers in the imitation and interactive conditions should be dissimilar to their own solo baseline, resulting in a decrease in intra-speaker similarity score.

\subsection{Data preparation}
\label{ssec:dataset}

To build the training and test datasets, we first segmented the audio files into phrase tokens with a Psychopy3 script based on timestamps. Non-voiced segments and noises like laughs and coughs were cut out, but stutters and repeats within the sentences were included. Stereo audios have been converted to mono audios. 39-dimensional MFCC features (13 static with $\Delta$ and $\Delta\Delta$) were extracted every \SI{10}{\milli\second} with a \SI{25}{\milli\second} window size using the Python package Librosa \cite{mcfee2015librosa}. Cepstral Mean and Variance Normalization were performed to mitigate the mismatch between the recording devices and the variation of the recording environment.

We used the solo data to train, validate and test the model and then further test the model performance on the interactive and imitation data. For each speaker, we created positively labelled data and negatively labelled data. A positive data example was a sentence pair produced by the same speaker, likewise the negative one from different speakers. We only used the real interactive speaker pairs to build the datasets. Surrogate speaker combinations have not been adopted.

The training set was constructed using the first 40 sentences of the script, the validation set using the following 20 sentences, and the test set using the remaining 20 sentences. Using this approach, we generated 45,240 positive data examples and 46,400 negative data examples for the training set. For the validation and test sets, we created 11,020 positive data examples and 11,600 negative data examples each. The data split ratio was thus approximately 70\%:15\%:15\%. The dataset comprises a balanced proportion of participants with diverse L1 backgrounds, and the label distribution is approximately equal across all groups.

To evaluate the model in interactive and imitation conditions, we constructed another test set by extracting a subset of sentence combinations. Positive samples were created using \textit{adjacent sentences} spoken by the same speaker, while negative samples were created using the \textit{same sentence} spoken by different speakers. This arrangement enables phonetic convergence and imitation measurement in Section \ref{sec:detection}. To build the interactive set, we conducted four rounds of ART experiments. Given the stable phonetic convergence observed in our previous work \cite{dejong22_interspeech}across four sessions, we randomly selected two sessions with different initial speakers.

\subsection{Training Details}
\label{ssec:training}

Our Siamese RNN model used 50 nodes in both the RNN and feedforward layers. 
The model trainable parameters included $\mathbf{W}_{h/y/e}$, $\mathbf{U}_h$, and $\mathbf{b}_{h/y/e}$, with a total size of 14,250 and were initialised
with small random normal-distributed values. 
Weight optimization was performed using the Adam \cite{kingma2014adam} method, together with a  learning rate decay strategy. Dropout (rate=0.2, after the RNN layer), $l_1$ regularization and batch-normalization were employed to avoid overfitting and stabilize the training process. The model's loss function was binary cross-entropy and the evaluation metrics were binary accuracy, precision, recall, F1-score and AUC. Our best result was achieved by initializing the model with weights pre-trained on the VCTK corpus \cite{Yamagishi2019vctk}, a large spoken sentence dataset with 109 L1 English speakers.

\subsection{Results}
\label{ssec:res}

Table \ref{table:1} summarises our experimentation on different model architectures, training data size and test scenarios which will be discussed in Section \ref{sec:discussion}. In Table \ref{table:2}, we report the precision, recall, F1-score, and AUC of our best-performing model for the binary speaker verification task. The first column shows the ART experimental condition under which the test was conducted. The results indicate that the model performs best in the solo condition, achieving an F1-score of 0.95 for the positive samples and 0.94 for the negative. The model's performance slightly drops in the interactive and imitation conditions as the speech characteristics are unconsciously or deliberately modified in the new scenarios.

\begin{table}[h]
\centering
\caption{Evaluation results of the best model}
\begin{tabular}{lcccc}
\toprule
&\textbf{Precision} & \textbf{Recall} & \textbf{F1-score} & \textbf{AUC}\\
\midrule
\textbf{Solo} & & & &\\
\midrule
positive & 0.93 & 0.97 & 0.95 & 0.99\\
negative & 0.96 & 0.92 & 0.94 & 0.99\\
\midrule
\textbf{Interactive} & & & &\\
\midrule
positive & 0.81 & 0.87 & 0.84 & 0.91\\
negative & 0.86 & 0.80 & 0.82 & 0.91\\
\midrule
\textbf{Imitation} & & & &\\
\midrule
positive & 0.82 & 0.86 & 0.84 & 0.91\\
negative & 0.85 & 0.81 & 0.83 & 0.91\\
\midrule
\bottomrule
\end{tabular}
\label{table:2}
\end{table}

\section{Phonetic convergence measurement}
\label{sec:detection}
We computed the average similarity scores of all speakers across three conditions with false predictions and outliers removed. Results show that either the intra-dyad similarity score of the imitation condition (0.081±0.092) or that of the interactive condition (0.074±0.085) is comparatively higher than in the solo condition (0.024±0.028). Meanwhile, the intra-speaker cosine similarity shows an opposite tendency with a gradual decrease from the solo condition (0.958±0.052), to the interactive (0.890±0.126) and the imitation condition (0.871±0.139). The results agree with our assumption about the dynamic of phonetic convergence across the experimental conditions. The intra-speaker similarity change is more significant than that of the dyadic setting.

\begin{figure}[h]
\centering
\includegraphics[width=0.47\textwidth]{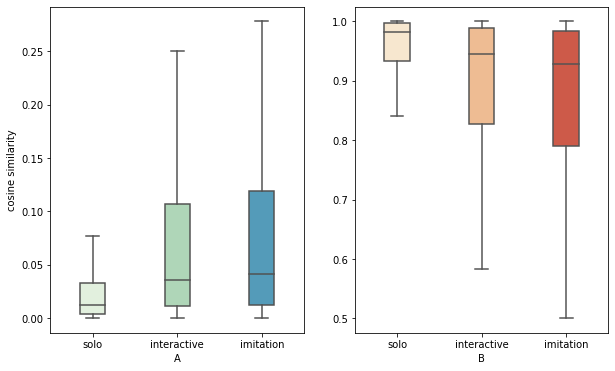}
\caption{The cosine similarity scores across the solo, interactive and imitation conditions. Subplot (A) shows intra-dyad similarity scores (different speakers) and (B) the intra-speaker similarity scores (same speaker).}
\label{fig:conv_dect}
\end{figure}
\begin{figure}[h]
\centering
\includegraphics[width=0.47\textwidth]{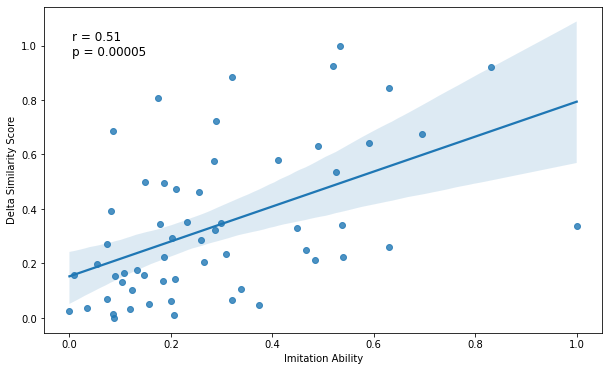}
\caption{The correlation between imitation ability and the degree of convergence in interaction with a 95\% confidence interval. The blue dots represent speakers. The imitation ability score on the x-axis and the cosine similarity change during interaction on the y-axis were normalized to (0, 1).}
\label{fig:imit_conv}
\end{figure}

We also tested whether a better speech imitator has a higher convergence degree in interaction. The imitation ability was defined by the average change of intra-speaker similarity across the solo and imitation conditions. The degree of phonetic convergence was measured as a speaker's average similarity change across the solo and interactive conditions. Figure \ref{fig:imit_conv} suggests a positive correlation between the imitation ability and the convergence degree (Pearson correlation coefficient r=0.51, p=0.0005) in the interactive condition. The convergence measurement result shown in Figure \ref{fig:conv_dect} and the correlation found in Figure \ref{fig:imit_conv} are consistent with our previous work\cite{dejong22_interspeech} using a GMM-UBM model in similar tasks. 

\section{Discussions}
\label{sec:discussion}
\textbf{A simple architecture with fine scalability.} The Siamese RNN model has shown its efficacy in speaker verification and convergence measurement tasks, yielding consistent results compared to the traditional GMM-UBM model, despite the notable model differences. The Siamese RNN model is characterized by its light-weighted structure and small parameter size, which facilitate its training and adaptation. We investigated more complex architectures such as LSTMs without pre-training, but they did not outperform the bi-directional RNN (Table\ref{table:1}a). Additionally, we evaluated the model's scalability with less training data. We built a subset using 10 sentences and retrained the model based on the pre-trained VCTK weights. The binary accuracy in 20 epochs reached 0.80 and increased to 0.87 in 100 epochs, as shown in Table \ref{table:1}b. Furthermore, we increased the number of speakers in the dataset and found that the model's performance did not decrease as discussed in the following paragraph.

\textbf{Adaptation to large speaker space to tackle L1-induced variability.} In our study, we observed a performance gap when testing the pre-trained VCTK model on L1 Italian, French, and Slovak data, indicating L1-induced data variability(Table \ref{table:1}c). To experiment on the model's ability to handle such variability, we randomly selected 5 female dyads from the Italian and French data pools to build training and validation sets. We trained two models, one with 5 additional random Italian and French female dyads and the other with 5 Slovak female dyads, and repeated the experiment five times to compare the models' average binary accuracy. Table \ref{table:1}d reveals that the model successfully captured the general acoustic features of the speakers. The introduction of new Slovak data did not confuse the model and its performance even improved.

\textbf{Subjective evaluation and feature engineering as next steps for model improvement.} The implementation of the Siamese architecture in this work has limitations as the interpretability of the phonetic convergence score is arguable due to the lack of ground truth. Therefore, subjective evaluation, such as AXB test\cite{pardo2006phonetic}, is necessary for future study of the model. Additionally, a good result was achieved with a relatively agnostic set of acoustic features such as MFCCs, the model's performance in the interactive and imitation condition could be improved by more refined feature engineering or the introduction of linguistic knowledge.

\section{Conclusions}
\label{ccl}
This paper presents a novel approach to studying phonetic convergence and speech imitation in L2-L2 interaction, utilizing a Siamese RNN model. The model was validated through a structured and controllable experiment, the scripted text alternated reading task (ART). In addition, the paper expanded the sample diversity by including Slovak L2 English speakers, which investigated the model's scalability and capability to handle L1-induced speaker variability.

\section{Acknowledgements}
\label{sec:tks}

This project has received funding from the European Union’s 
Horizon 2020 research and innovation programme under the 
Marie Skłodowska-Curie grant agreement No 859588.

\bibliographystyle{IEEEtran}
\bibliography{refs}

\end{document}